\documentclass[10pt,twocolumn,letterpaper]{article}

\usepackage{cvpr}
\usepackage{times}
\usepackage{epsfig}
\usepackage{graphicx}
\usepackage{amsmath}
\usepackage{amssymb}

\usepackage{amsmath}
\DeclareMathOperator*{\argmax}{arg\,max}
\usepackage{enumitem}
\usepackage{soul}
\usepackage{fancyhdr}

\graphicspath{ {images/} }

\usepackage[pagebackref=true,breaklinks=true,letterpaper=true,colorlinks=false,bookmarks=false]{hyperref}

\providecommand\K[1]{\textcolor{black}{#1}}
\providecommand\ENtwo[1]{\textcolor{black}{#1}}
\providecommand\cctwo[1]{\textcolor{black}{#1}}
\providecommand\ENquant[1]{\textcolor{black}{#1}}
\providecommand\ENthree[1]{\textcolor{black}{#1}}

\providecommand\KGCR[1]{\textcolor{black}{#1}}
\providecommand\KG[1]{\textcolor{black}{#1}}
\providecommand\KGtwo[1]{\textcolor{black}{#1}}
\providecommand\KGsupp[1]{\textcolor{black}{#1}}

\providecommand\KGthree[1]{\textcolor{black}{#1}}

\providecommand\cc[1]{\textcolor{black}{#1}}
\providecommand\EN[1]{\textcolor{black}{#1}}

\cvprfinalcopy 

\def\cvprPaperID{1792} 

\ifcvprfinal\fi
\begin{document}

\title{You2Me: Inferring Body Pose in Egocentric Video\\via First and Second Person Interactions}

\author{Evonne Ng$^{1,2}$\\
{\tt\small evonne\_ng@berkeley.edu}
\and
Donglai Xiang$^3$\\
{\tt\small donglaix@cs.cmu.edu}
\and
Hanbyul Joo$^4$\\
{\tt\small hjoo@fb.com}
\and 
Kristen Grauman$^{2,4}$\\
{\tt\small grauman@cs.utexas.edu}
\and
$^1$UC Berkeley
\and 
$^2$UT Austin
\and 
$^3$Carnegie Mellon University
\and
$^4$Facebook AI Research
}

\maketitle
\thispagestyle{fancy}
\fancyfoot{}
\chead{Appears in CVPR 2020}


\begin{abstract}
   The body pose of a person wearing a camera is of great interest for applications in augmented reality, healthcare, and robotics, yet much of the person's body is out of view for a typical wearable camera.  We propose a learning-based approach to estimate the camera wearer's 3D body pose from egocentric video sequences.  Our key insight is to leverage interactions with another person---whose body pose we \emph{can} directly observe---as a signal inherently linked to the body pose of the first-person subject.  
   We show that since interactions between individuals often induce a well-ordered series of back-and-forth responses, it is possible to learn a temporal model of the interlinked poses even though one party is largely out of view.  
   We demonstrate our idea on a variety of domains with dyadic interaction and show the substantial impact on egocentric body pose estimation, which improves the state of the art.
\end{abstract}

\section{Introduction}









Wearable cameras are becoming an increasingly viable platform for entertainment and productivity.  In augmented reality (AR), wearable headsets will let users blend useful information from the virtual world together with their real first-person visual experience to access information in a timely manner or interact with games.  
In healthcare, wearables can open up new forms of remote therapy for rehabilitating patients \cctwo{trying to improve their body's physical function in their own home}.  In robotics, wearables could simplify video-based learning from demonstration.

\begin{figure}[t]
\begin{center}
   \includegraphics[width=\linewidth]{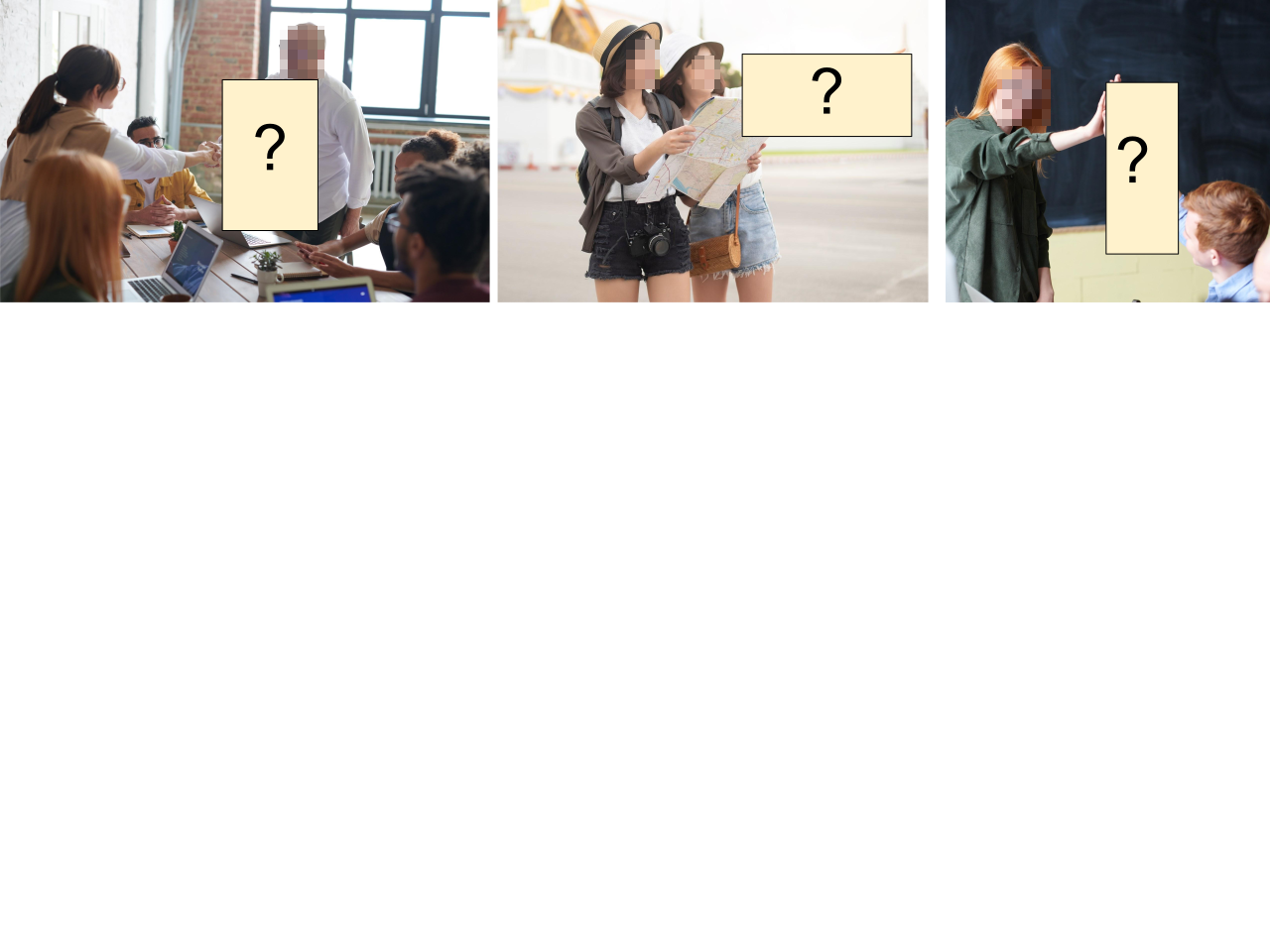}
\end{center}\vspace*{-1.8in}
   \caption{Inter-person interactions are common in daily activity \KG{and offer rich signals for perception.  Our work considers how interactions viewed from a first-person wearable camera can facilitate egocentric 3D body pose estimation.}}
   \vspace{-0.1in}
\label{fig:concept}
\label{fig:onecol}
\end{figure}

In all such cases and many more, the camera receives a first-person or ``egocentric" perspective of the surrounding visual world.  A vision system analyzing the egocentric video stream should not only extract high-level information about the visible surroundings (object, scenes, events), but also the current state of the person wearing the camera.  In particular, the \emph{body pose} of the camera wearer is of great interest, since it reveals his/her physical activity, postures, and gestures.  Unfortunately, the camera wearer's body is often largely out of the camera's field of view.  While this makes state-of-the-art third-person pose methods poorly suited~\cite{ref44,ref38,ref40,ref36,ref39,ref28}, recent work suggests that an ego-video stream nonetheless offers implicit cues for first-person body pose~\cite{ref1,ref17,kitani-iccv2019}. However, prior work restricts the task to static environments devoid of inter-person interactions, forcing the algorithms to rely on low-level cues like apparent camera motion or coarse scene layout.  
 
Our idea is to facilitate the recovery of 3D body pose for the camera wearer (or ``ego-pose" for short) by paying attention to the \emph{interactions} between the first and second person as observed in a first-person video stream.\footnote{\cctwo{Throughout,} we use ``second person" to refer to the person the camera wearer is \cctwo{currently} interacting with; if the wearer is ``I", the interactee  \cctwo{or partner in the interaction} is ``you".}  Inter-person interactions are extremely common and occupy a large part of any individual's day-to-day activities.
As is well-known in cognitive science~\cite{ramirez2011modeling, vinciarelli2012nonverbal, bernieri1988synchrony}, 
human body pose is \cctwo{largely} influenced by an inherent synchronization between interacting individuals. For instance, a person who sees someone reaching out their hand for a handshake will most likely respond by also reaching out \cctwo{their hand}; a person animatedly gesturing while telling a story may see their interacting partner nod in response; \cctwo{children playing may interact closely with their body motions.}
See Figure~\ref{fig:concept}.

\begin{figure}[t]
\begin{center}
   \includegraphics[width=\linewidth]{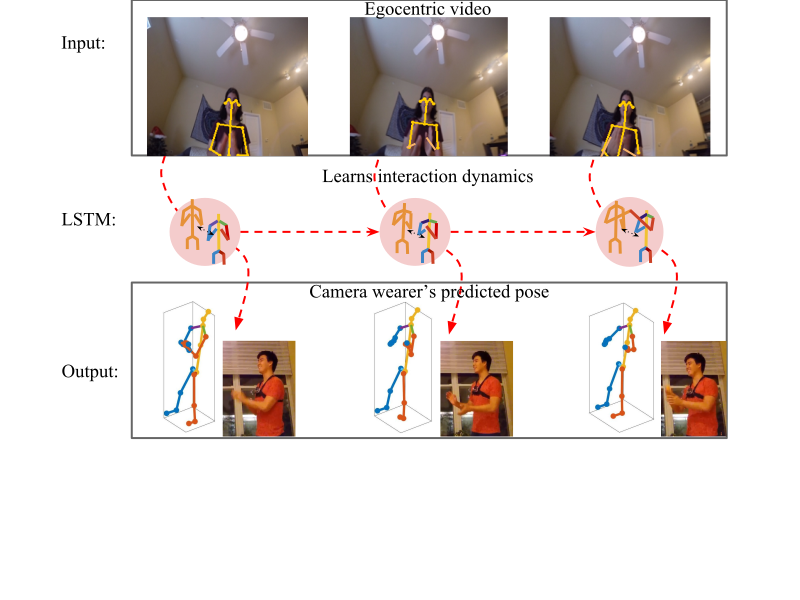}
\end{center}\vspace*{-0.80in}
   \caption{Our goal is to infer the full 3D body pose sequence of a person from their egocentric video captured by a single chest-mounted camera. 
   \K{Our model} focuses on 
   interaction dynamics to predict the wearer's pose 
   \K{as a function of the} interactee's pose, which is visible from the ego-view. The figure shows the input video with the interactee's (second-person) pose highlighted, and the output 3D joint predictions of the wearer's pose with corresponding pictures of the camera wearer. Note that our approach sees only the egocentric video (top); it does not see the bottom row of images showing the ``first person" behind the camera.}
   \vspace*{-0.1in}
\label{fig:concept2}
\label{fig:onecol}
\end{figure}

To that end, we introduce ``You2Me": an approach for ego-pose estimation that explicitly captures the interplay between the first and second person body poses. Our model uses a recurrent neural network to incorporate cues from the \emph{observed} second-person pose together with the camera motion and scene appearance to infer the \emph{latent} ego-pose across an entire video sequence. See Figure~\ref{fig:concept2}. 

\ENtwo{To our knowledge, no prior work models interactions for ego-pose. 
\K{Our} key contribution is to leverage the action-reaction dynamics in dyadic interactions to estimate 
\KGthree{ego-pose} from a monocular wearable video camera.}

We validate our You2Me ego-pose approach on two forms of ground-truth capture---from Kinect sensors and a Panoptic Studio \EN{\cite{ref75}}---on video data spanning \EN{10} subjects and several interaction domains (conversation, sports, hand games, and ball tossing).  Our results demonstrate that even though the first-person's body is largely out of view, the inferred second-person pose provides a useful prior on likely interactions, significantly boosting the estimates possible 
\K{from}
camera motion and scene context alone.  \cctwo{Furthermore, our} You2Me \cctwo{approach} outperforms the state-of-the-art \cctwo{approach} for ego-pose as well as a current standard deep third-person pose method when adapted to our setting.



\vspace*{-0.05in}
\section{Related work}

\paragraph{Third-person body pose and interactions}

There is extensive literature on human body pose estimation from the \KG{traditional} third-person viewpoint, where the person is entirely visible~\ENtwo{\cite{ref42, rogez2017lcr, xiaohan2015joint, cheron2015p}}.  
Recent approaches \KG{explore novel CNN-based methods,} 
which have substantially improved the detection of \emph{visible} body poses in images and video~\ENtwo{\cite{ref28, ref32, ref33, ref35, ref36, ref39, ref43, ref40, liu2019skeleton, lin2017recognizing, lao2018rolling}}. 
\KG{Our approach instead estimates the largely ``invisible" first-person pose.}
Multi-person pose tracking 
investigates structure in human motion and inter-person interactions 
to limit potential pose trajectories~\cite{ref23, ref28}. 
\KG{Beyond body pose,} there is a growing interest in modeling human-human interactions \EN{\cite{ref64, huang2014action, morency2010modeling}} to predict pedestrian trajectories \cite{ref51, ref52, ref54}, analyze social behavior and group activities \cite{ref52, ref53, ref61, ref62,huang2014action}, and understand human-object interactions \cite{ref59, ref63,hico}. 
\KG{Our method also capitalizes on the structure in inter-person interactions.}  However, whereas these existing methods assume that all people are fully within view of the camera, our approach addresses interactions between \KG{an individual in-view and an individual \emph{out-of-view}, i.e., the camera wearer.}



\vspace*{-0.15in}
\paragraph{Egocentric video}
Recent egocentric vision work focuses primarily on recognizing objects \cite{ref3}, activities~\cite{ref2,fathi-activity-2011,mccandless,ref9,ref11,ref12,ref14,ref73,ref65}, \KG{visible} hand and arm pose \cite{ref7,ref18,ref19,rogez-upperbody-2015,rhodin2016egocap}, \KG{eye gaze~\cite{ref13}, or anticipating future camera trajectories~\cite{hyun-soo-park-cvpr2016,ref74}.}
\KG{In contrast, we explore 3D pose estimation for the camera wearer's full body, and unlike any of the above, we show that the inferred body pose of another individual 
directly benefits the pose estimates. }




\vspace*{-0.15in}
\paragraph{First-person body pose from video}
Egocentric 3D full body pose estimation has received only limited attention~\cite{ref1,ref15,ref17}. 
\KG{The first attempt to the problem is the geometry-based} ``inside-out mocap" approach~\cite{ref15}, which uses structure from motion (SfM) to reconstruct the 3D location of 16 body mounted cameras placed on a person's joints. 
\KG{In contrast, we propose a learning-based solution, and it requires only a single chest-mounted camera, which makes it more suitable and comfortable for daily activity.}

\K{Limited recent work extracts ego-pose from}
monocular first-person video~\cite{ref1,ref17}.
The method in~\cite{ref1}
infers the poses of a camera wearer by 
\K{optimizing} an implicit motion graph 
\K{with} an array of \K{hand-crafted} cost functions, \cctwo{including a sit/stand classifier}. 
\K{In contrast}, we propose an end-to-end learning method, \cctwo{which 
learns from the full visual frame.}
The method in~\cite{ref17}
uses a humanoid simulator in a control-based approach to recover the sequence of actions affecting pose, \KG{and is evaluated quantitatively only on synthetic sequences.}
Whereas both prior learning-based methods focus on sweeping motions that induce notable camera movements (like bending, sitting, walking)
, our approach improves the prediction of upper-body joint locations during sequences when the camera 
\KGthree{has only subtle motions} (like handshakes and other conversational gestures).  
\KG{Furthermore, unlike~\cite{ref17}, our method does not require a simulator and does all its learning directly from video \cctwo{accompanied by ground truth ego-poses.}}
\KG{Most importantly, unlike any of the existing methods~\cite{ref1,ref15,ref17}, our approach} discovers the connection between the dynamics in inter-person interactions and egocentric body poses. 



\vspace*{-0.1in}
\paragraph{Social signals in first-person video}
\KG{Being person-centric by definition, first-person video is naturally a rich source of social information.  Prior work exploring social signals focuses} on detecting
social groups \cite{ref66, ref67, ref70} and \KG{mutual} gaze \cite{ref68, ref69} or shared gaze~\cite{social-saliency} from ego-video. 
\KG{More relevant to our work}, the activity recognition method of~\cite{ref65} uses 
paired egocentric videos to learn gestures and micro-actions in dyadic interactions.
That approach captures the correlations among inter-person actions (\eg, pointing, passing \K{item}) 
in two synchronized video clips to better classify them.
\KG{However, whereas~\cite{ref65} requires} two egocentric videos at test time, our approach relies only on a single ego-video.
While eliminating the second camera 
\KG{introduces new technical challenges} 
(since we cannot view both the action and response), 
it offers greater \K{flexibility}.
Furthermore, we infer body pose, whereas~\cite{ref65} classifies 
\K{actions.}


\section{Our approach}

\KG{The goal is to take a single first-person video as input, and estimate the camera wearer's 3D body pose sequence as output.  Our main insight 
is to leverage not only the appearance and motion evident in the first-person video, but also an estimate of the second-person's body poses.}

In this section, we present \KGCR{a} 
model that uses first- and \KG{second}-person features---both extracted from monocular egocentric video---to predict the 3D joints of the camera wearer. 
\KG{We first define the pose encoding (Sec~\ref{sec:problem}) \KGCR{and the} three inputs to our network (Sec~\ref{sec:feat1} to~\ref{sec:feat3}), followed by the recurrent long short-term memory (LSTM) network that uses them to make \KGCR{pose} 
predictions for a video (Sec~\ref{sec:lstm}).}


\vspace*{-0.05in}
\subsection{Problem formulation}\label{sec:problem}
\vspace*{-0.025in}

Given 
\EN{$N$ video frames} from a chest-mounted camera, we estimate a \EN{corresponding } sequence of \EN{$N$} 3D human poses. \KG{Each output pose} \EN{ $p_t  \in \mathbb {R}^{3J}$} {is a stick figure skeleton of 3D points consisting of \KG{$J$} joint positions for the predicted body pose of the camera wearer at frame $t$.}
Note that our goal is to infer \K{articulated} pose \KG{as opposed to} \K{recognizing an} \KGCR{action.}


Each predicted 3D body joint is positioned in a person-centric coordinate system with its origin at the \EN{camera on the wearer's chest}. The first axis is parallel to the ground and points towards the direction in which the wearer is facing. The second axis is parallel to the ground and lies along the same plane as the shoulder line. The third axis is perpendicular to the ground plane. \KG{To account for people of varying sizes,} we normalize each skeleton for scale based on the shoulder width of the individual.

\subsection{Dynamic first-person motion features}\label{sec:feat1}
\KG{Motion patterns observed from a first-person camera offer a strong scene-independent cue about the camera wearer's body articulations, despite the limbs themselves largely being out of the field of view.}
For example, a sudden drop in elevation can indicate movement towards a sitting posture, or a counterclockwise rotation can indicate shoulders tilting to the left. 

To capture these patterns, we construct scene-invariant dynamic features by extracting a sequence of homographies between each successive video frame, following~\cite{ref1}. 
While a homography is only strictly scene invariant when the camera is purely rotating, the egocentric camera translates very little between successive frames when the frame rate is high. 
These homographies facilitate generalization to novel environments, since the motion signals are \KGtwo{independent of the exact appearance of the scene.}

 \EN{We estimate the homography from flow correspondences by solving a homogeneous linear equation via SVD~\cite{hartley2003multiple}.}
 Each element in the resulting \EN{$3 \times 3$} homography matrix is then normalized by the top-left corner element. The stack of normalized homographies over a given duration is used to represent the global camera movement within the interval. For frame $f_t$ at timestep $t$ in a given video, the motion representation is constructed by calculating the homographies between successive frames within the interval $[f_{t-15}, f_t]$. We then vectorize the homographies and combine them into a 
 \EN{$m_t \in \mathbb {R}^{135}$ vector,}
 which represents a half-second interval of camera movements preceding frame $f_t$ \KG{(for 30 fps video)}.


\subsection{Static first-person scene features}\label{sec:feat2}

While the dynamic features reveal important \KG{cues} for sweeping actions that induce notable camera movements, such as running
\K{or} sitting,
they are more ambiguous for sequences with little motion in the egocentric video. To account for this, \KG{our second feature attends to the appearance of the surrounding scene.}
In everyday life, many static scene structures are heavily associated with certain poses. For example, if the camera wearer leans forward to touch his/her toes, the egocentric camera may see the floor; \cctwo{if the camera wearer stands while looking at a computer monitor, the egocentric camera will see a different image than if the camera wearer sits while looking at the same monitor.}  \KG{As with the dynamic features above, the surrounding scene provides cues about ego-pose without the camera wearer's body being visible.}

To obtain static first-person scene features, we use a ResNet-152 model pre-trained on ImageNet.  
Dropping the last fully connected layer on the pre-trained model, we treat the rest of the ResNet-152 as a fixed feature extractor for video frames. Given frame $f_t$, we run the image through the modified ResNet-152, which outputs 
\EN{$ s_t  \in \mathbb {R}^{2048}$.}
\KG{Whereas the ego-pose method of~\cite{ref1} relies on a standing vs.~sitting image classifier to capture static context, we find our full visual encoding of the scene contributes to more accurate pose learning.}  
\KG{Note that this feature by default also captures elements of the second-person pose; however, without extracting the pose explicitly it would be much more data inefficient to learn it simply from ResNet features, as we will see in results.}

\begin{figure}[t]
\begin{center}
   \includegraphics[width=\linewidth]{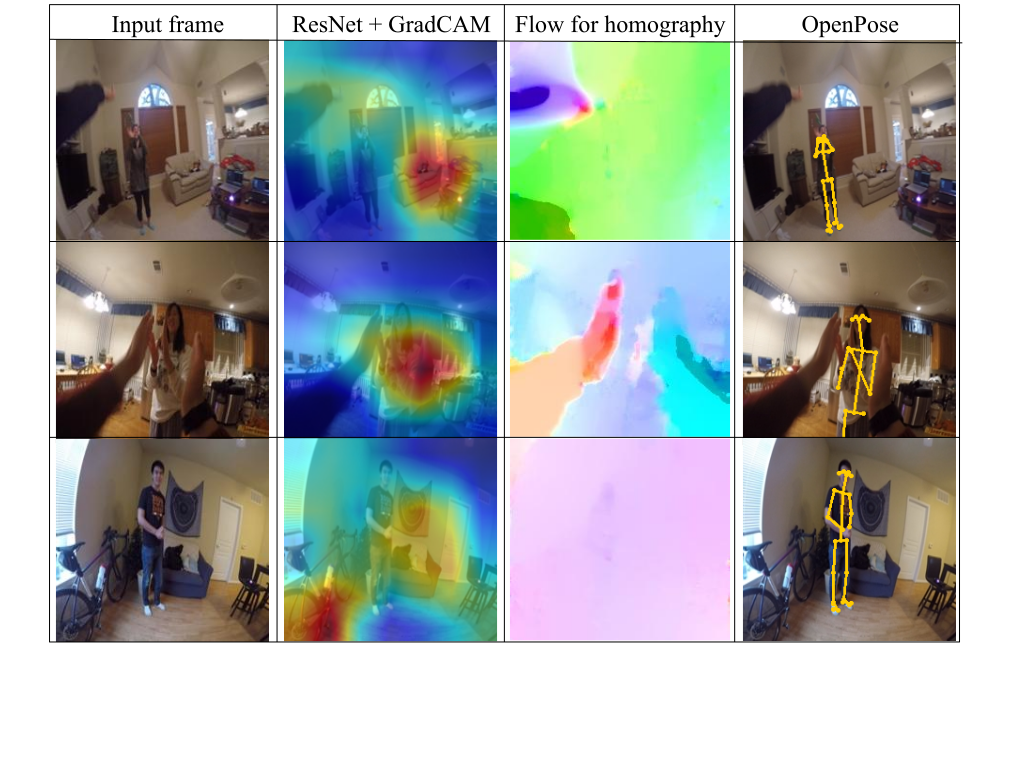}
\end{center}\vspace*{-0.55in}
   \caption{Visualization of features extracted from ego-video frames. The ResNet Grad-CAM~\cite{selvaraju2017grad} heatmaps suggest that when a person is further away, the focus is on static objects in the room (couch, bike, wall rug) \KGtwo{which help capture coarse posture}, but when the interactee is closer, the focus is more on the person, \KGtwo{which influences finer details}. While the flow/homography does especially well capturing motion from the camera wearer's hands, many sequences lack global motion and produce flows similar to the bottom row example. OpenPose~\cite{ref28} generates a 2D representation of the interactee's pose even with slight occlusions.}
   \vspace*{-0.05in}
\label{fig:features}
\label{fig:onecol}
\end{figure}

\subsection{Second-person body pose interaction features}\label{sec:feat3}

\KG{Our third and most important input consists of the ``second-person" pose of the person with whom the camera wearer is interacting.}
Whereas both the dynamic and static features help capture poses that come from larger common actions, we propose to incorporate second-person pose to explicitly account for the \emph{interaction dynamics} that influence gestures and micro-actions \KG{performed in sequence between two people engaged in an interaction.}

In human-human interactions, there is a great deal of symbiosis between both actors. Specific actions \KGCR{elicit} 
certain reactions, which in turn influence the body pose of the individual. For example, if we see an individual windup to throw a ball, our natural response is to raise our arms to catch 
the ball. \KG{Or more subtly, if we see a person turn 
to look at a passerby, we may turn to follow their gaze.}
By understanding this dynamic, we can gather important ego-pose information for the camera wearer by simply observing the visible pose of the person with whom he/she interacts. 

Thus, 
\KG{our third feature records the} interactee's inferred pose. 
Still using the egocentric video, we estimate the pose of the interactee in each frame.
\KGtwo{Here we can leverage recent} successes for pose estimation from a third-person perspective:  
\KG{unlike the camera wearer, the second person \emph{is} visible, i.e., the ego-camera footage gives a  third-person view of the interactee.}
\KG{
We use OpenPose~\cite{ref28} to infer interactee poses due to its efficiency and accuracy, though other third-person methods can also be employed.}
OpenPose provides real-time multi-person keypoint detection: given a stack of frames, it returns a corresponding stack of \ENtwo{normalized} 25 2D keypoint joint estimations \ENtwo{(see Supp.~file)}. For each frame $f_t$, we flatten the output 25 keypoint estimates into a vector 
\EN{$o_t \in \mathbb {R}^{50}$} \KGtwo{(denoted $o_t$ for ``other").}  \K{We set missing or occluded joints in $o_t$ to zeros, which can provide its own signal about the wearer's proximity to the interactee (e.g., no legs visible when he/she is closer).}


Note that our learning approach \KGtwo{is flexible to the exact encodings of the ego- and second-person poses.} 
\ENtwo{
\K{Using normalized} 2D keypoints for the second-person pose \K{is compatible} 
\K{with using} person-centric 3D coordinates for the ego-pose (\K{cf.~Sec.~\ref{sec:problem}});
\K{whereas it would be problematic for a purely geometric method relying on spatial registration, for a learning-based approach mixing 2D and 3D in this way is consistent.}
\K{Furthermore, while perfect 3D second-person poses would offer the most complete information, e.g., avoiding foreshortening,} we find that state-of-the-art 3D methods  
\cite{kanazawa2018end, xiang2019monocular, ref33} fail on our data due to extensive occlusions from the wearer's hands.}
\K{See Sec.~\ref{sec:results} for experiments that concretely justify this design choice.}

\KGtwo{Figure~\ref{fig:features} illustrates the complete set of \KGCR{features.}}



\begin{figure*}
\begin{center}
\includegraphics[width=0.9\linewidth]{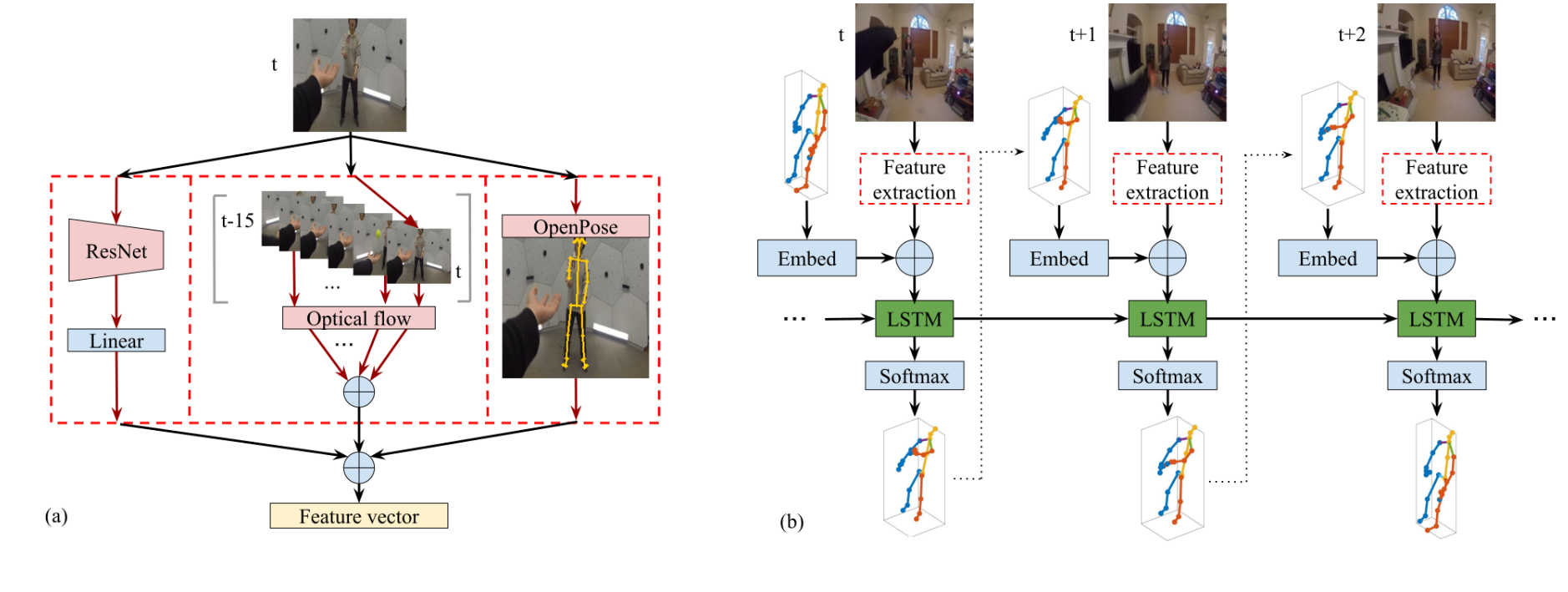}
\vspace*{-0.20in}
\end{center}
   \caption{Network architecture for our You2Me approach. (a) For each video frame, we extract three features. ResNet provides static visual cues about the scene.  
   \KGtwo{Stacked homographies for the past 15 frames provide motion cues for the ego-camera.}
   Finally, we extract the \KG{inferred} 2D pose of the visible interactee with OpenPose~\cite{ref28}. 
   All three \KG{features} are concatenated ($\oplus$) and fed into the LSTM. (b) illustrates our LSTM, which takes as input the feature vector from (a) and an embedding of the camera wearer's pose \KG{estimated from the previous frame}. Outputs from the LSTM produce ego-pose predictions, \KG{assigning one of the 
   \K{fine-grained} quantized body poses to each frame.}}
\label{fig:approach}
\end{figure*}

\subsection{Recurrent neural network for pose inference}\label{sec:lstm}

\KG{All three video-based cues defined above serve as input to a recurrent neural network to perform pose estimation for the full sequence.}
\KG{In particular, we define a Long Short-Term Memory (LSTM) network~\cite{graves2013generating, graves2014towards} for our task.}
The LSTM learns the current state of the camera wearer\KG{, scene, and interactee,} and uses this \KG{encoding} to \KGthree{infer} the \KG{camera wearer's} \KGthree{next} poses.   
The LSTM's hidden state captures the sequential patterns of linked body poses that result from inter-person back-and-forth responses.

\KG{While the LSTM can be trained to perform regression on the real-valued coordinates of the body pose, we found a \K{fine-grained} classification task to train more robustly, as often reported in the literature (\K{see} last row of Table \ref{tab:1}). Hence, we first quantize the space of training body poses into a large number of fine-grained poses using $K$-means \K{(details below)}.  Now the task is to map to the closest possible quantized pose at each time step.}

\EN{Given a hidden state dimension of $D$,} the hidden state \EN{vector $h_t \in \mathbb{R}^{D}$} of the LSTM at time $t$ captures the \KG{cumulative} latent representation of the camera wearer's pose at that  instant in the video. For each frame $f_t$, we extract the 
\EN{homography matrix $m_t$,}
the 
\EN{ResNet-152 scene feature vector $s_t$}, and the
\EN{second-person joint position vector $o_t$}. 
\KGtwo{To provide a more compact representation of the scene to the LSTM (useful to conserve GPU memory), we project $s_t$ to a lower-dimensional embedding $x_t \in \mathbb{R}^E$:}
\begin{equation}
    x_t = \phi_x(s_t; W_x),
\end{equation}
where $W_x$ \EN{is of size $E \times 2048$ and} consists of the embedding weights for $\phi_x(.)$. 
The embedding is then passed through a batch normalization layer.  

\EN{The LSTM uses the wearer's pose in the previous frame $p_{t-1}$ as input for the current frame. Let $p_{t-1}$ be a $K$-dimensional one-hot vector indicating the  pose for the camera wearer at the previous frame $t-1$ \KGCR{(initialized at $t=0$ as the average training pose)}.}
We learn a linear embedding for the pose indicator to map it to vector $z_t$:
\begin{equation}\label{pose_embed}
      z_t = \phi_z(\EN{p_{t-1}}; W_z),
\end{equation}
where $W_z$ \EN{is of size $E \times K$ and} consists of the learned embedding weights for $\phi_z(.)$.  

\KG{All the features} are concatenated (indicated by \KGthree{the} operation $\oplus$) into a single vector $b_t \EN{\in \mathbb {R}^{135+ 50+ 2E}}$:
\begin{equation}
b_t = m_t \oplus o_t \oplus x_t \oplus z_t,
\end{equation}
which is then used as input to the LSTM cell for the corresponding prediction at time $t$.  This introduces the following recurrence for the hidden state vector:
\begin{equation}\label{recursion}
    \begin{aligned}
        h_t &= \text{LSTM}(h_{t-1}, b_t; \theta_l),
    \end{aligned}
\end{equation}
\EN{where $\theta_l$ denotes the LSTM parameters.}

We define the loss for the network as the cross entropy loss across an entire sequence \KG{for predicting the correct (quantized) pose in each frame}.  Specifically, the loss $\mathcal{L}$ for a video of length \EN{$N$} is:
\begin{equation}\label{loss}
    \mathcal{L}(W_x, W_z, W_p, \theta_l) = \EN{ -\sum_t^N \log( \sigma_{P}(W_p h_t))},
\end{equation}
\KG{where $\sigma_{P}(\cdot)$ is the softmax probability of the correct pose ``class", and $W_p$ is the linear classifier layer of dimension $K \times D$.} 
Recall that the quantization is fine-grained, 
such that \K{each per-frame} estimate is quite specific; on average the nearest quantized pose \cctwo{in the codebook} is just \ENquant{0.27} 
cm away per joint (see Supp.~file).
\KG{The inferred pose ID at time $t$ (i.e., the $\argmax$ over the pose posteriors at that timestep) is taken as the input for $z_{t+1}$ for the subsequent frame.} 


\K{Rather than uniformly \KG{quantize} all $J$ joints, we perform a mixed granularity clustering to account for the more subtle pose changes concentrated on the upper body.}
\K{L}ower body poses across frames \K{exhibit less variance}, but upper body poses have more fine-grained important differences (\K{e.g., from arm gestures}).
\ENquant{Hence, we use a sparser $K$-means clustering ($K_{bot}=100$) for the lower body joints and denser ($K_{upp}=700$) for the upper body. 
Any given pose is thus coded in terms of a 2D cluster ID.  }

At test time, we use the trained LSTM to predict the 
\ENtwo{pose sequence.}
\EN{From time $t-1$ to $t$, we use the predicted cluster $\hat{p}_{t-1}$ from the previous LSTM cell in 
Eq. \ref{pose_embed}.}  
\KGtwo{Figure~\ref{fig:approach} overviews the model; see Supp.~for architecture details.}





\section{\KG{You2Me Video Datasets}}

We present a first-person interaction dataset consisting of  \EN{42} two-minute sequences from one-on-one interactions between 10  individuals. We ask each individual (in turn) to wear a chest mounted GoPro camera and perform various interactive activities with another individual. We collect their egocentric video 
then synchronize it with the body-pose ground truth for both the camera wearer and the individual standing in front of the camera. The dataset captures four classes of activities: \emph{hand games}, \emph{tossing and catching}, \emph{sports}, and \emph{conversation}. The classes are broad enough such that intra-class variation exists. For example, the sports category contains instances of (reenacted) basketball, tennis, boxing, etc.; \KG{the conversation category contains} 
\EN{individuals playing charades, selling a product, negotiating, etc.}  \KGtwo{In about 50\% of the frames, no first-person body parts are visible.}  \cc{To ensure that our approach is generalizable,} we employ two methods of capture, as detailed next.

\vspace*{-0.15in}
\paragraph{Panoptic Studio capture}
\KG{Our first capture mode uses a Panoptic Studio dome, following~\cite{ref75}.}
The studio capture consists of \EN{14} sequences recorded in $1920 \times 1080$ resolution at 30 fps using the GoPro Hero3 chest mounted camera on the medium field of view setting. The \KG{ground truth} skeletons of the camera wearer and the 
\K{interactee} 
are then reconstructed at 30 fps, matching the \K{ego-video} frame rate. 
Each skeleton is parameterized by $J=19$ 3D joint positions 
obtained using the method of~\cite{ref75}. \KG{Capturing video in the dome offers extremely accurate ground truth, at the expense of a more constrained background environment.}
A total of six participants of different height, body shape, and gender enacted sequences from each of the four activity classes.

\vspace*{-0.15in}
\paragraph{Kinect capture}
\KG{Our second capture mode uses Kinect sensors for ground truth poses.}
The Kinect capture consists of \EN{28} sequences also recorded in $1920 \EN{\times} 1080$ resolution at 30 fps. We use the GoPro Hero4 chest mounted camera on wide field of view setting. Both people's ground truth skeleton poses are captured at 30 fps using the Kinect V2 sensor. The pose is represented by $J=25$ 3D joint positions defined in the MS Kinect SDK. Given the greater mobility of the Kinect in contrast to the Panoptic Studio, we ask four participants to enact sequences from each of the activity classes in various places such as offices, labs, and apartment rooms. The videos from this dataset are taken in unconstrained environments but are all indoors due to limitations of the Kinect V2 sensor.  \KG{While Kinect-sensed ground truth poses are noisier than those captured in the Panoptic Studio, prior work demonstrates that overall the Kinect poses are 
well aligned with human judgment of pose~\cite{ref1}.}

\vspace*{0.05in}
We stress that our method uses \emph{only the egocentric camera video as input} at test time for both datasets.
Further, we emphasize that \KG{no existing dataset is suitable for our task.}
Existing pose detection and tracking datasets (\eg, \cite{ref71,  ref72})  are captured in the third-person viewpoint. \KG{Existing egocentric datasets are either limited to visible hands and arms~\cite{ref19,ref2},
contain only single-person sequences~\cite{ref1,ref71,ref72}, consist of synthetic test data~\cite{ref17}, or lack body-pose joint labels~\cite{ref65}.}
\KG{All our data \KGCR{is} publicly available.\footnote{\KGCR{\texttt{http://vision.cs.utexas.edu/projects/you2me/}}} See Supp video for examples.}

\section{Experiments}\label{sec:results}

We evaluate our approach on both the Panoptic Studio and Kinect captures. 
Our method is trained and tested in \KGthree{an} activity-agnostic setting: the training and test sets are split such that each set contains roughly an equal number of sequences from each activity domain (conversation, sports, etc.). For the Panoptic Studio, we train on \EN{7} sequences and test on \EN{7}.  
For the Kinect set, we train on \EN{18} sequences and test on \EN{10} that are recorded at locations not  seen in the training set. 
For both, we ensure that the people appearing in test clips do \emph{not} appear in the training set.


\vspace*{-0.1in}
\paragraph{Implementation details}
We generate training data by creating sliding windows of size 512 frames with an overlap of 32 frames for each sequence in the training set. \KGCR{This yields 3.2K training sequences and 2.3K test sequences.}
For the LSTM, we use an embedding dimension of \KGtwo{$E=256$}, 
fixed hidden state dimension of \KGtwo{$D=512$}, and  
batch size of 32. Learning rate is 0.001 for the first 10 epochs then decreased to 0.0001. 
\ENquant{In initial experiments, we found results relatively insensitive to values of $K_{upp}$ from 500 to 900 and $K_{bot}$ from 70 to 120, and fixed $K_{upp}=700$ and $K_{bot}=100$ for all results. \KGCR{Training time is 18 hours on a single GPU for 20 epochs; test time is 36 fps.}  See Supp.} 



\vspace*{-0.1in}
\paragraph{Baselines} 

\KG{We compare to the following methods:}
\vspace*{-0.08in}
\begin{itemize}[leftmargin=*]
    \item {\textbf{
    Ego-pose motion graph (\emph{MotionGraph})~\cite{ref1}}: 
the current state-of-the art method for predicting body pose from real egocentric video~\cite{ref1}.  
    We use the \KG{authors' code}\footnote{\texttt{http://www.hao-jiang.net/code/egopose/ego\_pose\_code.tar.gz}} and retrain their model on our dataset.  
    \KG{This method also outputs quantized poses; }\ENthree{we find their method performs best on our data for $K=500$.}
    
 \vspace*{-0.08in}
    \item {\textbf{Third-person pose deconv network (\emph{DeconvNet})~\cite{ref44}}}: We adapt 
    the human pose estimation baseline of~\cite{ref44} to our task.\footnote{\texttt{https://github.com/leoxiaobin/pose.pytorch}} Their approach adds deconvolutional layers to ResNet, and \KG{achieves the state-of-the-art on the 2017 COCO keypoint challenge.}
    We use the same network structure \cctwo{presented in the baseline}, but retrain it on our egocentric dataset\ENtwo{, altering the output space for 3D joints.}
    \KG{While this network is intended for detecting visible poses in third-person images, it is useful  to gauge how well an extremely effective off-the-shelf deep pose method can learn from ego-video.}
    
 \vspace*{-0.08in}
    \item {\textbf{Ours without pose information (\emph{Ours w/o $o_t$})}}: This is a simplified version of our model in which we do not feed the second-person \KG{2D} joints to the LSTM. \cctwo{The remaining network is unchanged and takes as input the extracted image features and homographies.}  \KG{This ablation isolates the impact of modeling interactee poses versus all remaining design choices in our method.}
    
 \vspace*{-0.08in}
    \item \textbf{Always standing (\emph{Stand})}} and {\textbf{Always sitting (\emph{Sit})}}: a simple guessing method (stronger than a truly random guess) that exploits the prior that most poses are somewhere near a standing or a sitting pose. The standing and sitting poses \KG{are} averaged over the training sequences.  
\end{itemize}


\vspace*{-0.2in}
\paragraph{Evaluation metric}
We rotate each skeleton so the shoulder is parallel to the yz plane and the body center is at the origin, then calculate error as the Euclidean distance between the predicted 3D joints and the ground truth, \KGtwo{averaged over the sequence and scaled to centimeters (cm) based on a reference shoulder distance of 30 cm.} \KG{Note that \K{we judge accuracy against the exact \emph{continuous ground truth} poses \ENthree{---\emph{not} whether we \KGthree{infer} the right pose cluster}.  While the predicted joints are a cluster center, the quantization is so fine-grained that on average the best discrete pose is only 0.70 cm from the continuous pose.}}


\begin{table}[t]
\begin{center}\footnotesize
\begin{tabular}{|l||c|c|c||c|c|c|}
\hline
& \multicolumn{3}{|c||}{Kinect} & \multicolumn{3}{c|}{Panoptic} \\
\hline
& \KG{Upp} & Bot & All & \KG{Upp} & Bot & All \\
\hline\hline
Ours & {\bf 15.3}  & {\bf 12.9}  & {\bf 14.3} & {\bf 6.5} & {\bf 12.0} & {\bf 8.6} \\
Ours w/o $o_t$ & 19.4 & 15.6 & 18.0 & 11.2 & 15.4 & 12.8 \\
MotionGraph~\cite{ref1} & 24.4 & 15.7 & 21.2 & 11.9 & 20.7 & 15.2 \\
DeconvNet~\cite{ref44} & 26.0 & 20.3 & 23.3 & 18.3 & 21.2 & 19.4 \\
Stand & 27.8 & 23.1 & 25.4 & 10.6 & 18.5 & 13.5 \\
Sit & 21.8 & 43.3 & 28.5 & 17.3 & 28.9 & 21.6 \\
\ENtwo{Ours as Regression} & 22.9 & 20.0 & 20.9 & 12.3 & 16.8 & 14.6 \\ 
\hline
\end{tabular}
\end{center}
\vspace*{-0.1in}
\caption{\EN{Average joint e}rror (cm) for all methods on the two dataset captures.   Our approach is stronger than the existing methods, and the second-person pose is crucial to its performance.}
\label{tab:1}
\end{table}


\begin{figure}[t]
\begin{center}
\includegraphics[width=\linewidth]{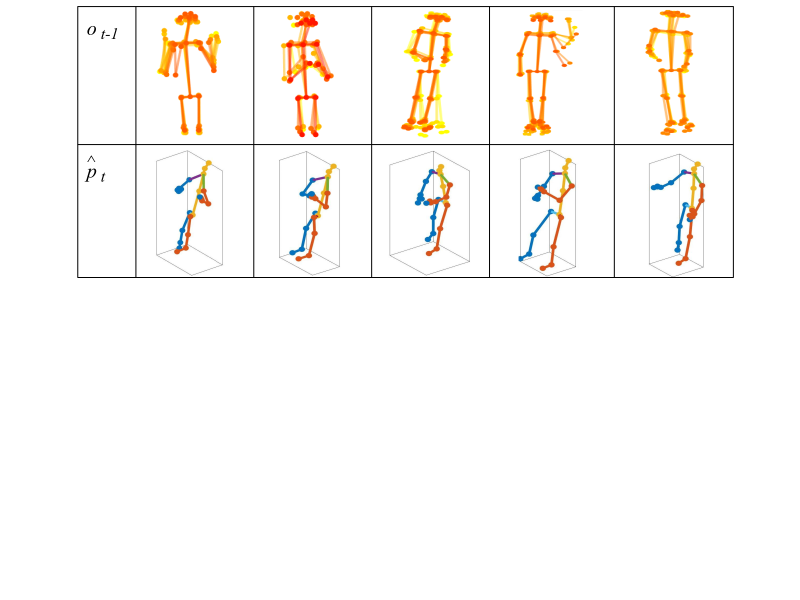}
\end{center}\vspace*{-1.48in}
   \caption{
   Most common second-person 2D poses (top) seen
   immediately preceding a given predicted 3D pose cluster (bottom) for test sequences.  You2Me captures useful interaction links like mutual reaches or tied conversation gestures.
   }
\label{fig:priors}
\label{fig:onecol}
\end{figure}

\begin{figure*}
\begin{center}
\includegraphics[width=0.8\linewidth]{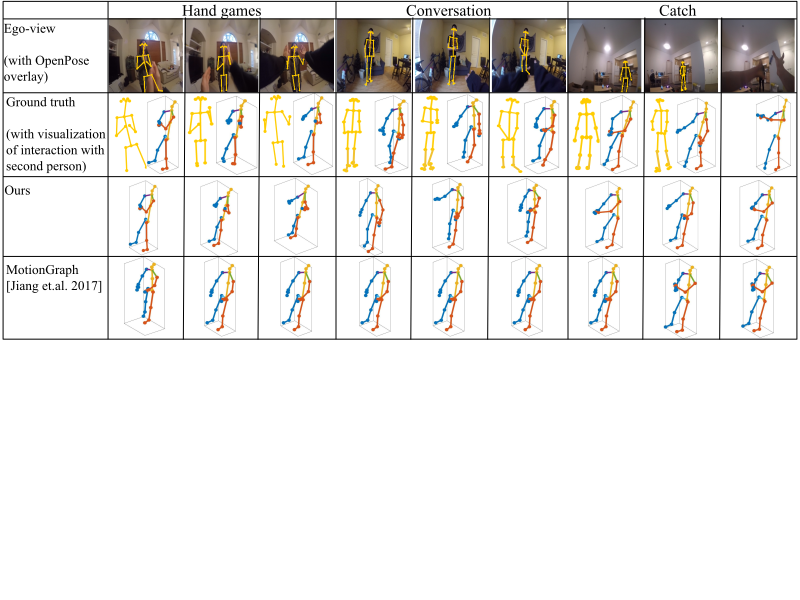}
\end{center}\vspace*{-1.95in}
   \caption{\KG{Example inferred poses for} three different activity domains trained in a domain-agnostic setting.  Row 1: ego-video view with OpenPose overlay \KG{(input to our method} is only the raw frame). Row 2: 3D ground truth poses \EN{in multicolor}, displayed as interacting with the 2D OpenPose skeletons in yellow. 
   Note: for ease of viewing, we show them side by side.  Row 3: results from our approach. Row 4: MotionGraph~\cite{ref1} results. In the last column, the interactee is fully occluded in the ego-view, but our predicted pose is still accurate.
  }
\label{fig:success}
\label{fig:onecol}
\end{figure*}

\begin{figure}[t]
\begin{center}
\includegraphics[width=\linewidth]{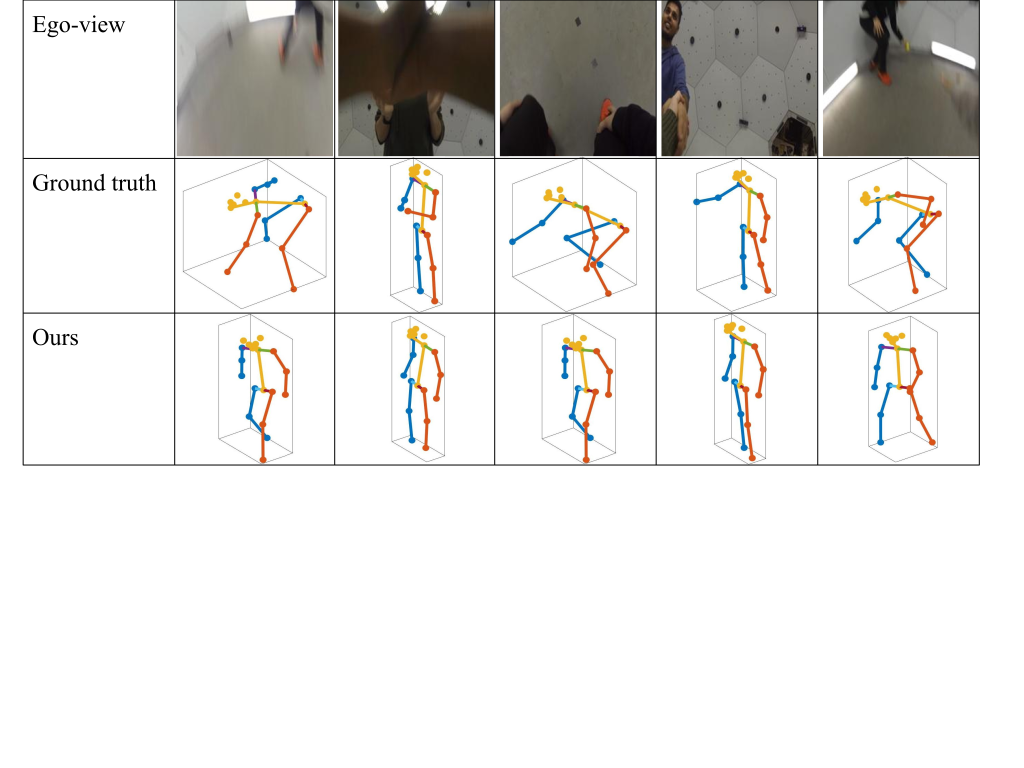}
\end{center}\vspace*{-1.13in}
   \caption{Example failure cases. 
   Typical failure cases are when the ego-view points at the ground or at feet, lacking the interactee's pose for a long duration. }
\label{fig:failure}
\label{fig:onecol}
\end{figure}

\vspace*{-0.1in}
\paragraph{Results}



Table \ref{tab:1} shows that the proposed method consistently gives better results than all of the competing methods. \KGtwo{We show errors averaged over all $J$ joints, and separately for the upper body joints which have the highest variance in everyday activity (head, elbow, wrists, hands) and the lower body joints (hips, knees, ankles, foot).} 
\ENtwo{See Supp.~file for \K{per-}joint errors.}
Our approach outperforms {\bf MotionGraph}~\cite{ref1} and {\bf Ours w/o $o_t$.} 
\KG{This result supports our key technical novelty of modeling mutual pose interactions between the first and second person.}
\KG{Our method's} improvement is even more significant in the upper body joints, \KG{which agrees with the fact that the most highly correlated inter-person poses occur with gestural motions of the head and arms.}   
The results show that the information provided by the pose of the interactee is essential for deriving 
\KG{accurate} body pose estimates for the camera wearer. 

We find our method's impact is greatest on conversation sequences, and lowest on sports sequences 
\ENtwo{(see Supp.~file).}
This suggests that during conversation sequences which involve less global motion, second-person pose provides essential information for more accurate upper body ego-pose predictions. In sports sequences, on the other hand, the interactee often moves out of view for long periods, 
explaining our method's lower degree of impact for sports.

While {\bf Sit} and {\bf Stand} offer a reasonable prior for most test frames, our method still makes significant gains on them, showing the ability to make more informed estimates on the limbs \ENquant{(e.g., 10 cm better on average} for the upper body keypoints). 
\ENtwo{{\bf Stand} outperforms \K{other} baselines \K{but not our method}. 
This is a well-known issue in motion forecasting: \cctwo{doing ``nothing" \K{can be} better than doing something, since} the average standing pose is in-between many test poses, representing a ``safe" estimate~\cite{martinez2017simple}. 
}
Our method also outperforms {\bf DeconvNet}~\cite{ref44}, which suggests that approaches for detecting poses from a third-person point of view 
\KG{do not easily adapt to handle the first-person pose task.} \K{Replacing our model's classification objective with regression is significantly weaker, supporting our design choice (see last row Table~\ref{tab:1}).}

\KGtwo{Figure~\ref{fig:priors} shows examples of the linked poses our method benefits from.  We display the second-person pose estimates immediately preceding various ego-pose estimates for cases where our method improves over the 
\textbf{Ours w/o $o_t$} baseline.  Intuitively, gains happen for interactions with good body language links, such as mutually extending hands or smaller conversational gestures.}

Figures~\ref{fig:success} and~\ref{fig:failure} show example success and failure cases for our approach, respectively. In Figure~\ref{fig:success}, our method outperforms \textbf{MotionGraph}~\cite{ref1}  in predicting upper body movements of the camera wearer, e.g., better capturing the swing of an arm before catching a ball
or reaching out to grab an object during a conversation. The failures in Figure~\ref{fig:failure} show the importance of the second-person pose to our approach. Analyzing the frames with the highest errors, we find failure cases occur primarily when the camera wearer is crouched over, the camera is pointed towards the floor, or the view of the interactee is obstructed. While our \KGthree{model} 
has enough priors to continue to accurately predict poses for a few frames without the interactee pose,  absent second person poses over extended periods are detrimental.


\begin{table}
\begin{center}\footnotesize
\begin{tabular}{|l||c|c|c||c|c|c|}
\hline
& \multicolumn{3}{|c||}{Kinect} & \multicolumn{3}{c|}{Panoptic} \\
\hline
& Upp & Bot & All & Upp & Bot & All \\
\hline\hline
Ours & {\bf 15.3} & {\bf 12.9} & {\bf 14.3} & {\bf 6.5} & {\bf 12.0} & {\bf 8.6} \\
w/o $x_t$ & 16.1 & 13.8 & 15.3 & 7.0 & 13.3 & 9.4 \\
w/o $o_t$ & 19.4 & 15.6 & 18.0 & 11.2 & 15.4 & 12.8 \\
w/o both & 20.0 & 16.9 & 18.8 & 10.2 & 15.3 & 12.1 \\
\hline
\end{tabular}
\end{center}\vspace*{-0.1in}
\caption{Ablation study 
\EN{to gauge the importance of the second-person pose features  $o_t$ and scene features $x_t$.}
Error in cm.}\label{tab:2}
\end{table}

Table \ref{tab:2} shows an ablation study, where
we add or remove features from our \KGthree{model} 
to quantify the impact of the second-person pose. \EN{Recall that  $o_t$ is the second-person pose and $x_t$ is the ResNet scene feature.} The results indicate that \textbf{Ours} and the \textbf{w/o $x_t$} model, which both use the \KG{second-person pose} (OpenPose estimates), consistently outperform the \textbf{w/o $o_t$} and \textbf{w/o both} models that lack the second-person pose estimate. Moreover, the results show that the addition of $o_t$ most significantly improves upper body predictions. 
The features of the interactee captured by the ResNet (\textbf{w/o $o_t$}) do not sufficiently capture the information encoded in the explicit pose estimate.

\begin{table}[t]
\begin{center}\footnotesize
\begin{tabular}{|l||c|c|c||c|c|c|}
\hline
& \multicolumn{3}{|c||}{Kinect} & \multicolumn{3}{c|}{Panoptic} \\
\hline
& Upp & Bot & All & Upp & Bot & All \\
\hline\hline
$o_t$ & 15.3 & 12.9 & 14.3 & 6.5 & 12.0 & 8.6 \\
GT \K{3D} & {\bf 13.8} & {\bf 12.3} & {\bf 13.2} & {\bf 6.0} & {\bf 8.7} & {\bf 6.9} \\
Still & 20.5 & 15.6 & 18.7 & 11.7 & 16.0 & 13.1 \\
Zero & 21.0 & 16.1 & 19.1 & 11.5 & 16.8 & 13.3 \\
Random & 22.4 & 16.6 & 20.4 & 11.8 & 17.7 & 14.3 \\
\ENtwo{Predicted 3D} & 19.3 & 15.2 & 17.8 & 11.0 & 16.8 & 13.0 \\
\hline
\end{tabular}
\end{center}
\vspace*{-0.1in}
\caption{Effects of second-person pose source.  Error in cm.}\label{tab:3}
\end{table}


Table~\ref{tab:3} analyzes \K{how the source of}  the second-person pose estimates affects our results.  First, we substitute in for $o_t$ the \K{3D} ground truth (GT) skeleton of the interactee, i.e., the true pose for the second person as given by the Panoptic Studio or Kinect.  We see that more accurate second-person poses can further improve results, \cctwo{though the margins are smaller than those separating our method from the baselines.} 
\K{Using 2D OpenPose for $o_t$ is better than using predicted 3D poses~\cite{ref33};} 3D pose from monocular data remains challenging. This justifies our use of 2D for $o_t$.
 \KGtwo{Next, to confirm our network properly learns a correlative function between the interactee pose and the ego-pose, we feed \emph{incorrect} values for $o_t$: either the average standing pose (Still), empty poses (Zero)}\EN{, or random poses from another sequence of another class 
(Random).} In all cases, the network produces poorer results, showing that our method is indeed leveraging the true structure in interactions.




\K{Please see Supp.~file for videos, per-joint and per-activity error breakdowns, depiction of the fine-grained quantization,} \ENthree{impact of quantization on approach, and additional architecture details.}

\section{Conclusions}

We presented \EN{the} You2Me approach to predict a camera wearer's pose given video from a single chest-mounted camera.  Our key insight is to capture the ties in interaction between the first (unobserved) and second (observed) person poses.  Our results
on two capture scenarios from several different activity domains demonstrate the promise of our idea, and we obtain state-of-the-art results for ego-pose.  

\KGtwo{Future work will include reasoning about the absence of second-person poses when interactions are not taking place, extending to sequences with multiple ``second people", and exploring how ego-pose estimates might reciprocate to boost second-person pose estimates.}

\noindent\textbf{Acknowledgements:} We thank Hao Jiang for helpful discussions. \KGCR{UT Austin is supported in part by ONR PECASE and NSF IIS-1514118.}



{\small
\bibliographystyle{ieee}
\bibliography{egbib}
}

\clearpage

{\Large{\textbf{Appendix}}}
\appendix
\section{\KG{Video example results}}
\setlength\tabcolsep{1.5pt}

The supplementary video\footnote{\texttt{http://vision.cs.utexas.edu/projects/you2me/demo.mp4}} shows video sequences of various test subjects and capture locations. 

We show examples of success cases across the four different action domains: \emph{conversation, sports, hand games, and ball tossing}. In both the Kinect and Panoptic Studio captures, our method is able to perform well. Most notably, our approach is able to determine when the camera wearer is going to squat or sit, when they are raising their hand to receive or catch an item, and when they are gesturing as part of a conversation. 

\KG{Consistent with the quantitative results provided in the paper,} compared against the baselines, we notice a significant difference between our approach and the \textbf{MotionGraph} \KG{[Jiang et al. CVPR 2017]}. While our approach is able to detect when the camera wearer is clapping as part of a hand game, the \textbf{MotionGraph} fails to do so. Similarly, when we remove the OpenPose features, \textbf{Ours w/o $o_t$} is also unable to capture when a person's hand is raised. However, our approach is even able to detect when the camera wearer is returning a single handed clap or a double handed clap in the hand-game.

For the failure cases, we demonstrate that our approach can fail when the view of the second person is obscured for a long duration. This is especially true for the sports category. Our method can also fail due to ambiguities in interaction dynamics. For example, in different hand games where clapping is common, there are many possible poses the camera wearer might be in given that the interactee has their hands clasped together. This is especially true towards the beginning of the sequence when there is little to no prior information.


\ENtwo{\section{Preprocessing of OpenPose 2D poses}
We normalize the 25 2D poses outputted by OpenPose by setting keypoint\_scale=3 (see manual),  
\ENthree{which scales the $(x,y)$ coordinates of the 2D poses to the range $[0,1]$, where $(0,0)$ is the top-left corner of the image and $(1,1)$ the bottom-right. }
We zero-out missing joints so the system learns proximity to the wearer (e.g., no legs if closer)}

\section{Pose clusters}

\begin{figure*}
\begin{center}
\includegraphics[width=\linewidth]{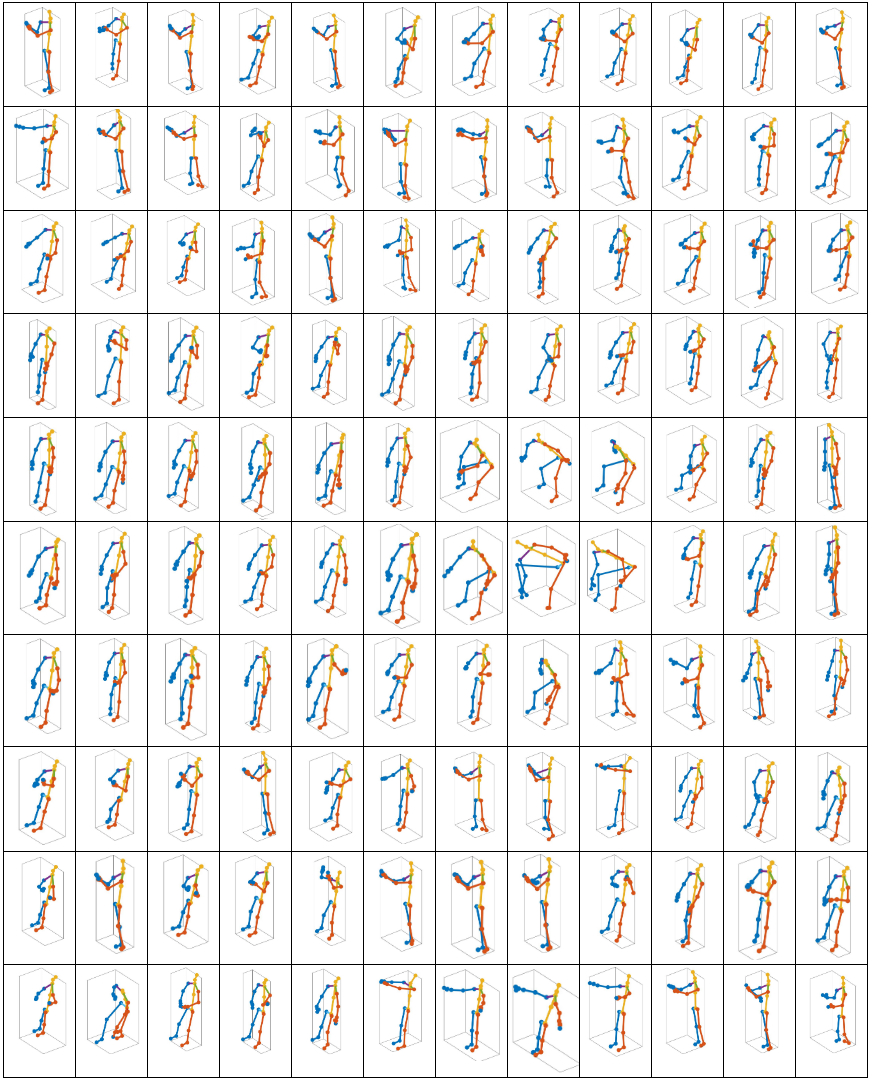}
\vspace*{-0.30in}
\end{center}
   \caption{120 randomly chosen pose clusters \KGsupp{(of the total 700 $\times$ 100 = 70,000 quantized poses)} \ENthree{shows the fine-grained granularity of the possible poses.}} 
\label{fig:cluster}
\end{figure*}

\KG{As noted in the main paper, the pose quantization is quite fine-grained, with average joint distances to the nearest pose of} \ENthree{0.27 cm}. 
We \KG{visualize} the granularity of differences between each cluster center in Figure \ref{fig:cluster}. 
With \ENthree{$K_{upp} = 700$ and $K_{bot} = 100$}
pose clusters, we get a good diversity of poses, enough to reasonably capture all possible poses in the training set. Additionally, the poses are fine-grained enough to accurately capture smaller movements of the arms and the legs (gesturing, micro-actions), or intermediate poses in larger movements (swinging, walking, sitting).

\section{Architecture details}
\ENthree{To implement the controlled quantization of upper and lower body poses, we train two networks: LSTM$_{upp}$ and LSTM$_{bot}$ to learn the upper and lower body poses, respectively. Both networks are set up identically to extract the same features and to output a pose cluster ID for each frame. The only difference is the granularity of the quantization. We found two LSTMs to be more effective than (1) treating each 2D cluster ID as a unique pose with a single LSTM (exceeds GPU memory) or (2) reducing the dimensionality by taking only the most frequent discrete poses (sacrifices precision of estimates). At each time frame, the full-body skeleton is then constructed by combining the predicted upper and lower body clusters.}

\EN{\KGsupp{We} train \ENthree{both} LSTMs with an embedding dimension of 256, a hidden state dimension of 512, and a batch size of 32. Additionally, we set the number of recurrent layers to 2. The LSTM\ENthree{s are} trained with a fixed sequence length of 512 for a total of 20 epochs. The learning rate for the first 10 epochs is 0.001 and is reduced to 0.0001 for the remaining 10 epochs. }

\EN{Furthermore, rather than feeding in raw video to the LSTMs, we first perform some preprocessing on the images. Each raw video is extracted at a frame rate of 30 fps. The frames are then resized to 224 x 224 x 3 images and normalized with a mean of $[0.485, 0.456, 0.406]$ and standard deviation of $[0.229, 0.24, 0.225]$ across the three channels. A stack of these preprocessed images serves as input to the LSTMs.}

\ENthree{\section{Impact of different quantizations on results}}
\ENthree{We provide further analysis of our method's controlled quantization. 
We first trained a single LSTM (\textbf{Single}) with an identical architecture setup as described above, but use full-body pose clusters as opposed to separating upper from lower. With a binary search on the cluster number, we achieve best results for this quantization method when $K=500$.}

\begin{table}
\begin{center}\scriptsize
\begin{tabular}{|l|c|c|c||c|c|c|}
\hline
& \multicolumn{3}{|c||}{Kinect} & \multicolumn{3}{c|}{Panoptic} \\
\hline
& Upp & Bot & All & Upp & Bot & All \\
\hline\hline
Ours & {\bf 15.3}  & {\bf 12.9}  & {\bf 14.3} & {\bf 6.5} & {\bf 12.0} & {\bf 8.6} \\
Ours w/o $o_t$ & 19.4 & 15.6 & 18.0 & 11.2 & 15.4 & 12.8 \\
Single & 17.0 & 14.9 & 15.5 & 10.2 & 14.7 & 11.9 \\
Single w/o $o_t$ & 25.7 & 18.9 & 22.0 & 16.8 & 20.5 & 18.2 \\
MotionGraph [26] & 24.4 & 15.7 & 21.2 & 11.9 & 20.7 & 15.2 \\
MotionGraph2 [26] & 27.4 & 14.8 & 22.9 & 13.9 & 22.0 & 16.9 \\ 
\hline
\end{tabular}
\end{center}
\vspace*{-0.1in}
\caption{\ENthree{Average joint error (cm) for all methods on the two dataset captures. Comparison of quantization methods shows while more controlled quantization improves results, the second-person pose is crucial to its performance.}} 
\vspace*{-0.12in}
\label{tab:originalLSTM}
\end{table}

\ENthree{As shown in Table \ref{tab:originalLSTM}, having the more controlled quantization improves performance. However, note that \textbf{Single} already outperforms \textbf{MotionGraph} [Jiang et al. CVPR 2017], which uses an identical full-body quantization with $K=500$, and \textbf{MotionGraph2}, which is a modification of \textbf{MotionGraph} [Jiang et al. CVPR 2017] using the \KGsupp{same} controlled quantization \KGsupp{as our model} ($K_{upp}=700$ and $K_{low}=100$).   \KGsupp{In short, our model outperforms \textbf{MotionGraph} in every scenario, no matter if either method uses the same quantization from that paper or the proposed quantization.}
Furthermore, \textbf{Single} outperforms \textbf{Ours w/o $o_t$}, and \textbf{Single w/o $o_t$}, both of which lack the second-person feature. Thus while the controlled quantization helps improve results, the second-person pose alone provides significant improvements over state-of-the-art.}

\ENthree{\section{Convolutional autoencoder details}}
\ENthree{For the Supp video, we train a convolutional autoencoder [Holden et al.~SIGGRAPH 2015] to smooth the pose trajectories produced by the LSTM. However, please note all evaluation \KGsupp{in the main paper} uses the \emph{unsmoothed} LSTM outputs. The fully convolutional network learns to produce non-quantized results from the quantized LSTM outputs. This post-processing is used to remove jitter and to generate more realistic motion for visualization.}

\ENthree{To construct the training set, we generate quantized pose sequences by mapping ground truth poses to their nearest cluster. We then divide both the quantized pose sequences and the original ground truth sequences into fixed sized chunks $F = 265$, such that each chunk is $F \times J$, where $J$ is the skeleton dimension. The quantized sequence serves as input to the autoencoder while the ground truth is the target output. We standardize the input and output skeletons using the mean and variation of the training set. To create the Supp video, we apply the autoencoder on the unsmoothed output of the LSTM to get final smoothed results. }

\begin{table}[t]
\begin{center}\scriptsize
\begin{tabular}{|l|c|c|c|c|c|c|}
\hline
 & should & elbows & hands & hips & knees & feet \\
\hline \hline
\ENthree{Ours} & \textbf{4.1 (3.0)} & \textbf{8.5 (6.6)} & \textbf{15.9 (9.0)} & \textbf{5.3 (5.6)} & \textbf{10.1 (9.3)} & \textbf{16.6 (15.2)} \\
\ENthree{Ours w/o $o_t$} & 4.6 (3.9) & 13.6 (13.7) & 28.1 (18.3) & 7.5 (6.9) & 13.5 (13.3) & 20.7 (21.3) \\
MotionGraph [26] & 5.6 (5.1) & 15.2 (10.6) & 28.5 (18.6) & 7.0 (6.8) & 14.1 (21.2) & 19.9 (29.9) \\
\hline
\end{tabular}
\end{center}
\vspace*{-0.1in}
\caption{Breakdown of error (cm) for notable joints in interactions, Kinect/(Panoptic).  \KGsupp{This table shows the per joint errors for the same experiment reported in summary form in Table 1 of the main paper.}} 
\label{tab:perjoint}
\end{table}

\begin{table}[t]
\begin{center}\scriptsize
\begin{tabular}{|l||c|c|c||c|c|c|}
\hline
& \multicolumn{3}{|c||}{Ours} & \multicolumn{3}{c|}{Ours w/o $o_t$} \\
\hline
& Upp & Bot & All & Upp & Bot & All \\
\hline\hline
\ENthree{Conversation} & 14.8 (6.0) &  14.2 (12.9) & 14.4 (8.3) & 22.4 (12.5) & 18.2 (17.0) & 20.8 (13.8) \\
\ENthree{Hand Games} & 15.6 (6.5) & 12.3 (12.2) & 14.3 (8.3) & 21.7 (12.2) & 13.7 (16.4) & 18.8 (13.4)\\
\ENthree{Throw-Catch} & 10.4 (5.1) & 7.5 (7.1) & 9.2 (5.7) & 15.2 (9.5) & 9.8 (10.6) & 13.2 (9.7) \\
\ENthree{Sports} & 13.5 (9.8) & 20.9 (19.2) & 15.7 (12.9) & 14.9 (12.8) & 22.2 (20.5) & 17.1 (15.2) \\
\hline
\end{tabular}
\end{center}
\vspace*{-0.1in}
\caption{Breakdown of joint error (cm) per activity, Kinect/(Panoptic).  \KGsupp{This table shows the results from the same experiment reported in summary form in Table 2 of the main paper.}}
\vspace*{-0.12in}
\label{tab:per_action}
\end{table}

\ENtwo{\section{Per joint and per activity analysis} 
In the main paper, we provide analysis for: 1) errors vs.~state-of-the-art baselines [26,62] on two distinct datasets (Table 1), 2) influence of activity type (L700-16), 3) example influential 2nd-person poses (Fig 5), 4) successes/failures vs.~[26] (Fig 6,7,video), 5) ablation study (Table 2), 6) impact of errors in 2nd-person pose (Table 3).  Throughout, we report separate errors for upper/lower joints.} 

\ENtwo{To give a more fine-grain analysis of our results, we further break down upper body errors (``Upp" in Table 1 of our submission) per arms and hand joints in Table~\ref{tab:perjoint}. Again, we make significant gains for joints relevant to interactions (L693-99). Errors for [26] are \ENthree{at least} $\sim$50\% higher than ours on hands/elbows on \ENthree{both datasets.} 
}

\ENtwo{Furthermore, we analyze the per activity errors for Ours vs.~Ours w/o $o_t$ to evaluate which activities are more or less influenced by additional 2nd-person pose information, as discussed in L700-16 in the main paper. Table~\ref{tab:per_action} further breaks down the results from Table 2 \ENthree{in the main paper} and shows that 2nd-person pose helps even for non-symmetric cases (like throw-and-catch).}

\end{document}